\documentclass[conference]{IEEEtran}
\IEEEoverridecommandlockouts
\usepackage{cite}
\usepackage{romannum}
\usepackage[ruled, lined, linesnumbered, commentsnumbered, longend]{algorithm2e}
\SetCommentSty{mycommfont}
\usepackage{xcolor}
\usepackage{tikz}
\usetikzlibrary{shapes.geometric, arrows}
\usepackage{multirow}
\usepackage{tabu}
\usepackage{amsmath,amssymb,amsfonts}
\usepackage{algorithmic}
\usepackage{graphicx}
\usepackage{textcomp}
\usepackage{subfigure}
\usepackage{xcolor}

\usepackage{tabularx,booktabs}
\usepackage{adjustbox}
\newcolumntype{C}{>{\centering\arraybackslash}X} 
\usepackage{lipsum}

\def\BibTeX{{\rm B\kern-.05em{\sc i\kern-.025em b}\kern-.08em
    T\kern-.1667em\lower.7ex\hbox{E}\kern-.125emX}}

\SetAlgoSkip{}

\begin{document}

\title{Unsupervised Learning based Element Resource Allocation for Reconfigurable Intelligent Surfaces in mmWave Network
\\
}

\author{\IEEEauthorblockN{\textsuperscript{1}Pujitha Mamillapalli, \textsuperscript{2}Yoghitha Ramamoorthi, 
\textsuperscript{1}Abhinav Kumar,\\
\textsuperscript{2}Tomoki Murakami, \textsuperscript{2}Tomoaki Ogawa  and \textsuperscript{2}Yasushi Takatori} 
\IEEEauthorblockA{\textit{\textsuperscript{1}Department of Artificial Intelligence, Indian Institute of Technology Hyderabad, Telangana, India}}
\IEEEauthorblockA{\textit{\textsuperscript{2}NTT Access Network Service Systems Laboratories, Nippon Telegraph and Telephone Corporation, Yokosuka, Japan}}
}






\maketitle

\begin{abstract}

The increasing demand for high data rates and seamless connectivity in wireless systems has sparked significant interest in reconfigurable intelligent surfaces (RIS) and artificial intelligence-based wireless applications. RIS typically comprises passive reflective antenna elements that control the wireless propagation environment by adequately tuning the phase of the reflective elements. The allocation of RIS elements to multiple-user equipment (UEs) is crucial for efficiently utilizing RIS. In this work, we formulate a joint optimization problem that optimizes the RIS phase configuration and resource allocation under an $\alpha$-fair scheduling framework and propose an efficient way of allocating RIS elements. Conventional iterative optimization methods, however, suffer from exponentially increasing computational complexity as the number of RIS elements increases and also complicate the generation of training labels for supervised learning. To overcome these challenges, we propose a five-layer fully connected neural network (FNN) combined with a preprocessing technique to significantly reduce input dimensionality, lower computational complexity, and enhance scalability. The simulation results show that our proposed NN-based solution reduces computational overhead while significantly improving system throughput by $6.8\%$ compared to existing RIS element allocation schemes. Furthermore, the proposed system achieves better performance while reducing computational complexity, making it significantly more scalable than the iterative optimization algorithms.



\end{abstract}

\vspace{-5mm}
\section{Introduction}
The 5G and Beyond 5G networks are expected to use millimeter wave (mmWave) \cite{NTT} for a high data rate. However, the range of such networks is limited to shorter distances due to high attenuation and blockages. Reconfigurable intelligent surfaces (RIS) \cite{NTT} offers a promising way to boost coverage and data throughput in mmWave networks by adjusting the electromagnetic properties of signals. 

The allocation of RIS elements to multiple user equipment (UEs) plays a critical role in fully realizing the benefits of RIS. Uniform-sized RIS element allocation, where continuous groups of equal number of RIS elements are assigned to each UE, has been considered in~\cite{2b} and~\cite{3b} for reducing computational complexity and energy consumption, respectively. The optimal number of RIS elements required to maintain quality of service (QoS) has been investigated in~\cite{4b}. In \cite{NTT_ref}, RIS elements were dynamically allocated among multiple UEs based on their communication demand and distance from the RIS. Furthermore, a double deep policy gradient-based deep reinforcement learning (DRL) framework for uniform-sized allocation was explored in [5]. While effective, such DRL-based methods demand high computational resources, extensive hardware support, and long training times, which limits their feasibility for real-time and large-scale RIS-aided mmWave networks.

Conventional iterative optimization methods suffer from computational scalability issues as the number of RIS elements increases. These approaches often involve solving non-convex or combinatorial optimization problems, where the search space grows exponentially with RIS size~\cite{NP_hard}, making them impractical for large-scale deployments. For example, discrete phase-shift optimization for RIS is shown to be NP-hard, requiring exhaustive or near-exhaustive search when the number of elements becomes large~\cite{NPHard}. Furthermore, iterative algorithms used for RIS beamforming and resource allocation typically exhibit polynomial or higher-order complexity, which escalates rapidly with system dimensions~\cite{Drawbacks_IA}. In addition to scalability, such methods complicate the generation of training labels for supervised learning frameworks, since solving large-scale optimization problems repeatedly for different channel realizations and system configurations is highly resource-intensive and time-consuming.

To address these challenges, we propose a novel unsupervised learning-based neural network (NN) framework for RIS element allocation. Unlike conventional approaches, the proposed method avoids reliance on labeled data and mitigates the prohibitive computational cost associated with iterative optimization, thereby enabling scalable and efficient RIS allocation while preserving strong system performance. Existing studies that enforce uniform and contiguous RIS element allocation simplify the problem but at the expense of reduced system efficiency. To the best of our knowledge, this work is the first to introduce a discrete and non-uniform RIS element allocation scheme with fairness using an unsupervised NN-based solution, offering a low-complexity yet highly effective approach for practical large-scale RIS-assisted networks. With these motivations, the following contributions are made.

With these motivations, the following contributions are made. First, we formulate a joint optimization problem to maximize $\alpha$-fair throughput under the assumption of perfect channel state information (CSI), enabling a flexible trade-off between network efficiency and user fairness. To ensure computational feasibility, we relax the RIS constraints and reformulate the problem into a convex optimization framework. This reformulated problem is then efficiently solved using the block coordinate descent (BCD) method, which iteratively optimizes different variable blocks while guaranteeing overall convergence. Furthermore, we propose a five-layer fully connected neural network (FNN) combined with preprocessing technique as an alternative solution to significantly reduce input dimensionality, lower computational complexity, and enhance scalability. Finally, we conduct a comprehensive performance evaluation of the proposed framework and benchmark its performance against state-of-the-art schemes.


The organization of the paper is as follows. Section II explains the system model. Section III discusses the joint problem formulation of the proposed discrete RIS elemnet allocation. The architecture of unsupervised learning-based FNN is described in Section IV, followed by experimental data analysis in Section V. Section VI presents the conclusion and future work.

Notation: a, $\mathbf{a}$ and $\mathbf{A}$ denote scalar, column vector, and matrix, respectively. $(\cdot)^T$ , and $(\cdot)^H$ denote transposition, and conjugate, respectively. $\mathbf{A}(k,i)$ denotes the entry in the $k^{th}$ row and $i^{th}$ column of $\mathbf{A}$. ${\mathbb C^{n \times n}}$ represents an $n \times n$ matrix in the complex set. $\text{Real}(\cdot)$ and $\text{Imag}(\cdot)$ extract the real and imaginary parts of the argument, respectively. $\lVert{ \cdot }\rVert_{2}$ denotes the $l_{2}$ norm, and $j$ represents the imaginary unit.  $\nabla f(x)$ denotes the Euclidean gradient of the function $f(x)$ with respect to $x$ .

\begin{figure}[t]
\centerline{\includegraphics[scale=0.35]{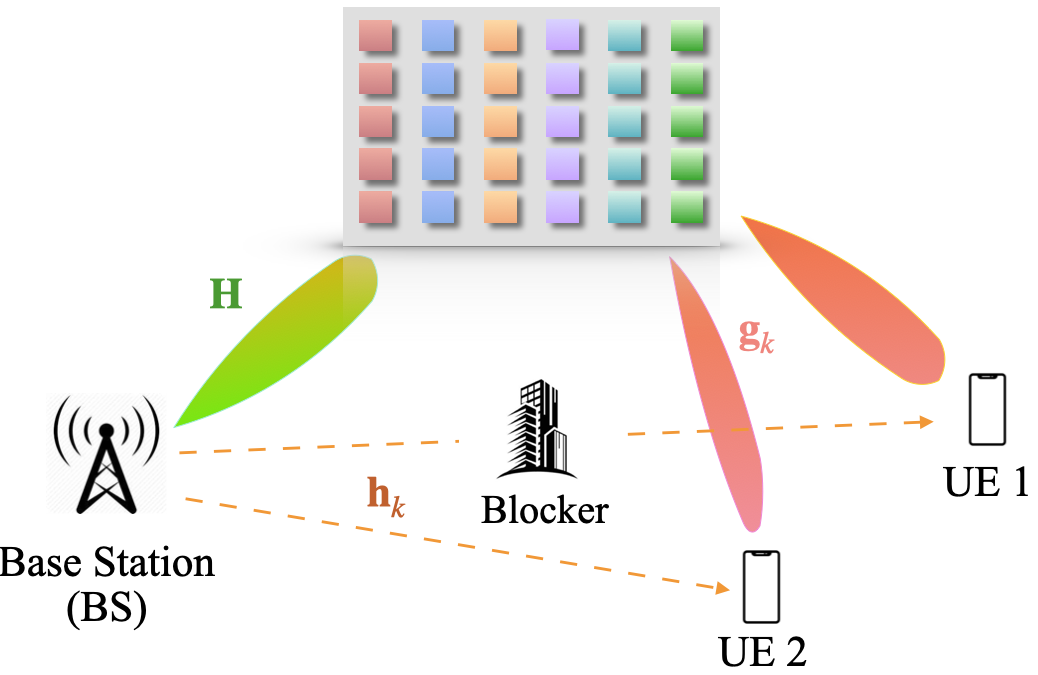}}
\vspace{-0.05in}
\caption{System model for RIS-aided mmWave network}
\label{System Model}
\vspace{-0.1in}
\end{figure}

\begin{table}[t]
\caption{Mathematical Notations}
\begin{center}
\begin{tabular}{|c|c|}
\hline
$N$ & Number of antennas in BS\\
\hline
$\lambda_k$ & Distribution density of UEs\\
\hline
$\lambda_b$ & Distribution density of blockages\\
\hline
$L^2$ & Total number of RIS elements \\
\hline

$L_s$& number of uniform RIS elemnets\\

\hline
$\mathbf{h}_k$ & Channel gain between BS and $k^\text{th}$ UE \\
\hline
$\mathbf{H}_{RB}$ & Channel gain between BS and RIS\\
\hline
$\mathbf{g}_k$ & Channel gain between RIS and $k^\text{th}$ UE\\
\hline
$\theta_l$ & phase shift of the $l^{th}$ RIS element\\
\hline
$\mathbf{\Xi}_k$ & binary allocation variable of RIS elemnets to $k^{th}$ UE\\
\hline
$P_t$ & Transmitted power at BS\\
\hline
$\mathbf{w}_k$ & active beamforming vector for the $k^{th}$ UE\\
\hline
$s_k$ & Baseband symbols for the $k^{th}$ UE\\
\hline
$y_k$ & Received signal at $k^\text{th}$ UE\\

\hline

$n_k$ & Additive white Gaussian noise (AWGN) with variance $\sigma^2$ \\
\hline
$I_k$ & interference from other UEs at the $k^{th}$ UE\\
\hline

$\gamma_k$ & Signal-to-interference-plus-noise ratio of the $k^{th}$ UE\\
\hline

$R_k$ & Achievable data rate of the $k^{th}$ UE\\
\hline

$d_\text{3D}$ & 3D-distance between the Tx and Rx\\
\hline
$d_\text{2D}$ & 2D-distance between the Tx and Rx\\
\hline
$d_\text{BP}^{'}$ & Break point distance\\
\hline
$h_\text{BS}$ & antenna height of the BS\\
\hline
$h_\text{UE}$ & antenna height of the UE\\
\hline
$f_c$ & center frequency\\
\hline
$\sigma_\text{SF}$ & log-normal shadow fading term\\
\hline
$\mathbb{U}_{\alpha }$ & $\alpha$-fair utility function\\
\hline
\end{tabular}
\label{notations}
\end{center}
\end{table}

\section{System Model}
We consider a downlink (DL) multiple-input-multiple-output (MIMO) system with a base station (BS) equipped with $N$ transmit antennas and an RIS with $L\times L$ antennas serving $K$ single-antenna UEs as illustrated in Fig.1. Let  $\mathcal{K}  = \{1, \ldots, K\}$ denote the set of UE indices, which are distributed based on an independent Poisson point process (PPP) with intensity $\lambda_k$. In addition, we consider blockages to be distributed based on PPP with the intensity of $\lambda_B$.

These blockages average length and width are $E[ln]$ and $E[wd]$, respectively. The presence of the blockage is determined based on the probability of blockage as in \cite{blockage}. If a blockage is detected, then attenuation due to blockage is computed as per the 3GPP documentation \cite{3GPP}. The received signal at $k^{th}$ UE in the DL scenario is defined as
\begin{align} 
y_{r,k} & = \left( {{{\mathbf{g}_k}^{H}}\mathbf{\Theta} {\mathbf{{H}}_{RB}} + {\mathbf{h}_k}} \right)\mathbf{w}_k s_k + {I}_k + {n_k}, ~ \forall k \in {\mathcal{K}},
 \label{eq1}
\end{align}
 where $\mathbf{w}_k$ is the corresponding beamforming vector at BS and $s_k$ is the baseband symbol for the $k^{th}$ UE. IN addition, ${I}_k = {\sum _{i=1, i \not= k}^{K}\left( {{{\mathbf{g}_i}^{H}}\mathbf{\Theta} {\mathbf{H}_{RB}} + {\mathbf{h}_i}} \right)\mathbf{w}_k s_k}$ is the interference from other UEs, $n_k\sim {\mathcal{ C}}{\mathcal{ N}}( {0,{\sigma_n ^{2}}} )$ is the additive white Gaussian noise. The channel between the BS and RIS is denoted by  $\mathbf{H}_{RB} \in \mathbb {C}^{L^2\times N}$, the channel between the RIS and the $k^{th}$ UE is denoted by $\mathbf{g}_k \in \mathbb {C}^{ L^2 \times 1}$, and the channel between the BS and the $k^{th}$ UE is denoted by $\mathbf{h}_k \in \mathbb {C}^{1 \times N}$. The phase of RIS elements is defined as $\mathbf{\Theta} =  \left[ {{e^{j{\theta _1}}},\cdots,{e^{j{\theta_l}}}, \cdots ,{e^{j{\theta _{L^2}}}}} \right] \in \mathcal{R}^{1 \times L^2}$, where $\theta_l$ represents the phase of the $l^{th}$ RIS element for $l = \{1, 2,…, L^2\}$, respectively.   
The channel between BS and RIS is defined as \cite{2b}
\begin{equation}
    \mathbf{H}_{RB}= \beta_{H} \mathbf{a}_T(\psi^e_{BS},\psi^a_{BS})  \mathbf{a}_R^H(\phi^e_{RIS},\phi^a_{RIS}) \in \mathbb {C}^{L^2 \times N},
\end{equation}
where $\beta_\mathbf{H}$ is the path loss component between BS and RIS, $\mathbf{a_T}$ and $\mathbf{a}_R$ are the response of the array that captures the beamforming gain at the transmitter and receiver, respectively. Here , ${\phi^e_{RIS}}$ and ${\phi^a_{RIS}}$ are the angle of arrival (AoA) in  elevation and azimuth at the RIS, respectively; ${\psi^e_{BS}}$ and ${\psi^a_{BS}}$ are the angle of departure (AoD) in elevation and azimuth at the BS, respectively. 
The channel between RIS and UE is defined as\cite{2b} 
\begin{equation}
    \mathbf{g}_k= \beta_{\text{g},k} \mathbf{a}_T(\psi^e_{RIS},\psi^a_{RIS}) \in \mathbb {C}^{L^2 \times 1},
\end{equation}
where $\beta_{\text{g},k}$ is the path loss component between RIS and $k^{th}$ UE, ${\psi^e_{RIS}}$ and ${\psi^a_{RIS}}$ are the AoD in elevation and azimuth at the RIS, respectively. In addition, the channel between BS and UEs is expressed as \cite{2b}
\begin{equation}
    \mathbf{h}_k=\beta_{\text{h},k} \mathbf{a}_T(\psi^e_{BU},\psi^a_{BU}) \in \mathbb {C}^{1 \times N},
\end{equation}
where $\beta_{\text{h},k}$ is the pathloss component between BS and $k^{th}$ UE, ${\psi^e_{BU}}$ and ${\psi^a_{BU}}$ are the AoD in elevation and azimuth at the BS to UE, respectively. The pathloss are calculate using the 3GPP documentation \cite{3GPP} with small scale fading loss as $\sigma_\text{SF}$. Although our system model does not explicitly account for mobility, these effects can still be incorporated using channel estimates available at the transmitter. 

\subsubsection{Outdoor Pathloss Model}
Reference path loss models \cite{3GPP} for line-of-sight (LOS) and non line-of-sight (NLOS) cases in the outdoor UMi environment are given as follows, respectively:

\begin{equation}
    PL_\text{UMi-LOS}=
    \begin{cases}
        PL_1 & \quad 10\,\text{m} \leq d_{\text{2D}} \leq d_{\text{BP}}^{\prime} \\
        \text{$PL_2$} & \quad \text{$d_\text{BP}^{\prime} \leq d_\text{2D}\leq$ 5km }
    \end{cases}
\end{equation}

\begin{equation}
    PL_1=32.4 + 21\log_{10}(d_\text{3D})+20\log_{10}(f_c) + \sigma_\text{SF}
\end{equation}


\begin{equation}
\begin{split}
 PL_2 &=32.4 + 40\log_{10}(d_\text{3D})+20\log_{10}(f_c)\\
    & -9.5\log_{10}((d_\text{BP}^{'})^{2} + (h_\text{BS}-h_\text{UE})^{2}) + \sigma_\text{SF} 
\end{split}
\end{equation}

\begin{equation}
    PL_\text{UMi-NLOS}= \max(PL_\text{UMi-LOS},PL_\text{UMi-NLOS}^{'})
\end{equation}

\begin{equation}
\begin{split}
PL_\text{UMi-NLOS}^{'} & =35.3\log_{10}(d_\text{3D}) + 22.4 + 21.3 \log_{10}(f_c) \\
& \quad - 0.3 (h_\text{UE}-1.5) + \sigma_\text{SF}
\end{split}
\end{equation}

\begin{itemize}
    \item Breakpoint distance $d_\text{BP}^{'}=4 (h_{Tx}-h_E) (h_{Rx}-h_E) f_c/c$, where $f_c$ is the frequency, $c$ is the propagation velocity in free space, $h_{Tx}$ is the height of the transmitter antennas, $h_{Rx}$ is the height of the reciever antennas and $h_E = 1.0 \text{m}$ for UMi scenario.
    \item $\sigma_\text{SF}$ is the shadow fading term, $d_\text{2D}$ and $d_\text{3D}$.
\end{itemize}

\subsection{RIS elemnet Allocation}
\begin{figure}[t]

    \centering
    \subfigure[]
    {\includegraphics[width=1.30in]{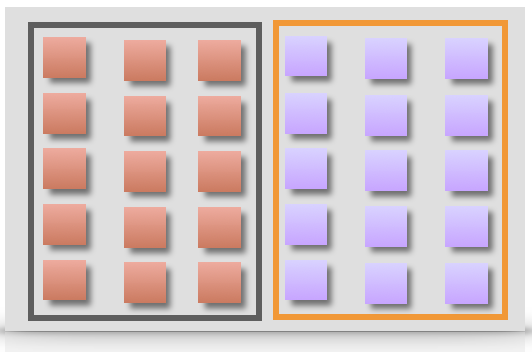}
       \label{Scene1}
    }
    \subfigure[]
    {\includegraphics[width=1.30in]{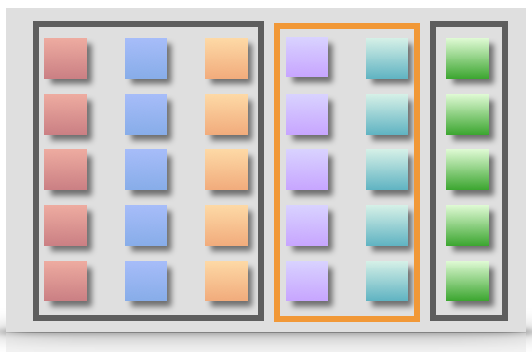}
     \label{scene2}
    }
    \caption{Two scenarios of RIS elemnet allocation (a) Existing RIS elemnet allocation: RIS elemnets are uniform sized continuous in nature and (b) Proposed RIS elemnet allocation: RIS elemnets are non-uniform sized discrete in nature}
    \label{scene}
\end{figure}

In contrast to uniform-sized RIS elemnet allocation \cite{3b}, where each subarray has a fixed size, we propose a more dynamic allocation strategy that allows for flexible subarray sizes and optimizing resource distribution based on UE requirements. The key differences between these approaches are illustrated in detail in Fig. \ref{scene}.
 \begin{figure*}[t]
\centerline{\includegraphics[scale=0.545]{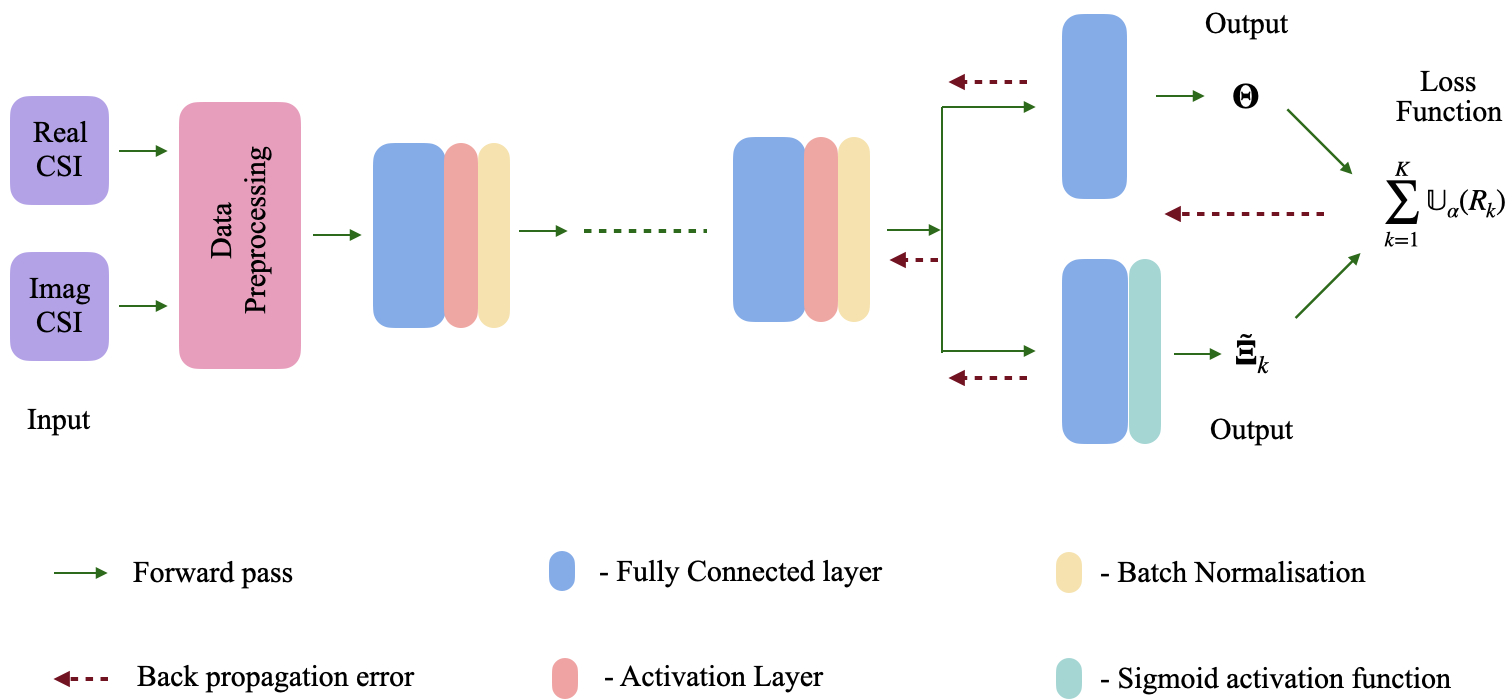}}
\centering\caption{Architecture of the fully connected neural network}
\label{CNN_archi}
\end{figure*} 
%
 
The RIS element allocation variable $\mathbf{\Xi}_k$ for the $k^{th}$ UE is then defined as
\begin{equation}
   \mathbf{\Xi}_k= [\xi_{k,1},\cdots,\xi_{k,l},\cdots,\xi_{k, {L}^2}] \in \mathbb{B}^{1 \times {L}^2},
   \label{xi}
\end{equation}
where $\xi_{k,l}=1$, whenever the $l^{th}$ RIS column is assigned to the $k^{th}$ UE; else 0.
collectively form the RIS elemnet dedicated to that UE as shown in Fig. \ref{scene2}. After the RIS elemnet allocation, the received signal at $k^{th}$ UE in (\ref{eq1}) is modified as:
\begin{align} &\hspace {-1pc}
    \hat{y}_{r,k} = \left( {{{\mathbf{{g}}_k}^{H}} \operatorname{diag} (\mathbf{\Xi}_k^H{\mathbf{\Theta}}) {\mathbf{{H}}_{RB}} + {\mathbf{h}_k}} \right)\mathbf{w}_k s_k \nonumber\\
    & + {I}_k + n_k, ~\forall k \in {\mathcal{K}},
     \label{eq2}
\end{align}

The signal-to-interference-plus-noise ratio ($\gamma_k$) of the $k^{th}$ UE is expressed \cite{2b} as
\begin{equation}
    \gamma_k= \frac{ \Big\lVert \left( {{{\mathbf{{g}}_k}^{H}} \operatorname{diag} (\mathbf{\Xi}_k^H{\mathbf{\Theta}}) {\mathbf{{H}}_{RB}} + {\mathbf{h}_k}} \right)\mathbf{w_k}\Big\rVert_2^2}{\Big\lVert {I}_k \rVert_{2}^2+\sigma_n^2},
    \label{SINR}
\end{equation}
Using (8) and (9), the normalized achievable data rate ($R_k$) of the $k^{th}$ UE as in \cite{2b} is written as follows: 
\begin{equation}
        R_k = \frac{1}{K}\log_2 \Bigg(1+{\frac{ \Big\lVert \left( {{{\mathbf{{g}}_k}^{H}}\operatorname{diag} (\mathbf{\Xi}_{k}^{H}{\mathbf{\Theta}}) {\mathbf{{H}}_{RB}} + {\mathbf{h}_k}} \right)\mathbf{w_k}\Big\rVert_2^2}{\lVert{I}_k\rVert_{2}^2+\sigma_n^2}}\Bigg),    
    \label{rate}  
\end{equation}
 which further can be expressed as a function of  ${\mathbf{\Theta}},\mathbf{\Xi}_k$ and the normalization factor 1/K in the rate $R_k$ is due to the loss in multiplexing loss when serving $K$ UEs. The utility function for an $\alpha$-fair scheduler with respect to the rate of  $k^{th}$ UE ($R_k$) is expressed as \cite{fairness}
\begin{equation} 
\mathbb {U}_{\alpha }(R_k)=\begin{cases} \dfrac {R_k ^{1-\alpha }}{1-\alpha }, & \text {$\alpha >0, \,\, \alpha \neq 1$,} \\ log(R_k), & \text {$\alpha =1$,} \end{cases} 
\label{utfn}
\end{equation}
 
\subsection{Performance Metrics}
In this work, $\alpha$-mean throughput is considered as the performance metric, which is expressed as follows \cite{fairness}
\begin{align} {T}_{\alpha }=\begin{cases} \displaystyle \left ({\frac {1}{|K|}\sum \limits _{k =1}^K (B * R_{k})^{1-\alpha } }\right)^{\frac {1}{1-\alpha }},&\text {$\alpha >0, \,\,\alpha \neq 1$,} \\ \displaystyle \left ({\prod \limits _{k =1}^K B * R_{k}}\right)^{\frac {1}{|K|}},&\text {$\alpha =1$,} \end{cases} \label{throughput_fn}\end{align}
where $\alpha$ is the fairness parameter, $B$ is the bandwidth, $K$ is the total number of UEs and $R_k$ is the rate as in (\ref{rate}). 

\section{Proposed Discrete RIS elemnet Allocation}
In this section, we propose a novel way of discrete RIS elemnet allocation shown in Fig. \ref{scene2}. These RIS elemnets are nonuniform in size and discrete in nature. 

Our objective is to maximize the sum of the $\alpha$-mean throughput of each UE with respect to the phase of RIS ${\mathbf{\Theta}}$ and binary allocation variable $\mathbf{\Xi}_k (\forall k \in \mathcal{K})$ as in (\ref{xi}), respectively. This formulation ensures fairness across users while capturing the trade-off between system-wide throughput and individual user performance through the adjustment of the parameter $\alpha$. Given the channel parameters $\{\mathbf{h}_k,{\mathbf{g}}_k,{\mathbf{H}}_{RB}\}_{k=1}^K$,  we formulate the resource allocation problem as below
\begin{eqnarray}        
   \text {\textbf{P0}}: & &  \max \limits _{\boldsymbol{{\Theta},\mathbf{\Xi}_k}}\quad \sum \limits ^{K}_{k=1} \mathbb {U}_{\alpha }(R_k), \label{eq:obj_fn} \\
     s.t. & & R_k \triangleq   f({\mathbf{\Theta}},\mathbf{\Xi}_k) \quad \text{as in (\ref{rate})} \label{eq:constraint1} \nonumber \\
    & &  \sum_{k=1}^{K} \sum_l^{{L}} \xi_{k,l} \leq {L}  \label{eq:constraint2},\\
    & &  \sum_{k=1}^{K}  \xi_{k,l} \leq 1, ~\forall l\in \mathcal{L} \label{eq:constraint3},\\
    & &  0\leq {\theta}_{l} \leq \pi, ~\forall l\in \mathcal{L}  \label{eq:constraint4},\\
    & &  \xi_{k,l}  \in \{0,1\}, ~ \forall k \in \mathcal{K} ~\& ~ l\in \mathcal{L}  \label{eq:constraint5}, 
\end{eqnarray}   
where $R_k$ is the rate of the $k^\text{th}$ UE as mentioned in  (\ref{rate}).  The constraint (\ref{eq:constraint2}) ensures that the sum of RIS columns allocated for all UEs is less than equal to the total available columns, and (\ref{eq:constraint3}) ensures that each column is allocated to a single user. (\ref{eq:constraint4}) ensures that the phase of RIS is within the angle $[0,\pi]$. The constraint (\ref{eq:constraint5}) represents the binary nature of the allocation variable.
A possible way of optimizing problem \textbf{\textbf{P0}} is using the brute-force approach. Then, the computational complexity will be $O(\nu^L KL)$; where $\nu$ is the number of all possible angles for the phase induced by the RIS elements. Therefore, computational complexity, which is in the order of the number of RIS columns, is a massive hindrance in both simulated and real-time applications.
\begin{algorithm}[t]
    \SetKwFunction{isOddNumber}{isOddNumber}
    \SetKwInOut{KwIn}{Input}
    \SetKwInOut{KwInt}{Initialize}
    \SetKwInOut{KwR}{Repeat}
    \SetKwInOut{KwOut}{Output}
     \SetKwInOut{KwU}{Until}
    \KwIn{$\mathbf{X}=\{\mathbf{h}_k,{\mathbf{g}}_k,{\mathbf{H}_{RB}}\}_{k=1}^K$}
    \KwInt{Initialize the optimization variables: $\mathbf{{\Theta}}^{(0)}$, $\tilde{\mathbf{\Xi}}_k^{(0)}$}
    \textbf{Repeat}
        $i \leftarrow i+1$
        
        Update $\mathbf{{\Theta}}^{(i)}$, $\tilde{\mathbf{\Xi}}_k^{(i)}$ using CVX solver
        
        Compute $f({\mathbf{\Theta}},\tilde{{\mathbf{\Xi}}}_k)$ for the given input $\mathbf{X}$  
        
    \textbf{Until} Convergence
    \caption{BCD Algorithm}
    \label{BCD}
\end{algorithm}
\vspace{-0.85mm}

The problem \textbf{P0} is a non-convex optimization problem, as the constraint (\ref{eq:constraint5}) is a non-convex constraint due to its discrete nature. The nonconvexity of the problem \textbf{P0} is mainly due to the presence of binary variables $\xi_{k,l}$ makes it NP-hard. To address this problem, we relax $\xi_{k,l}$ to obtain $\tilde{\xi}_{k,l}$ , where $\xi_{k,l}\in \{0,1\}$ is replaced by $\tilde{\xi}_{k,l} \in [0,1], \forall$ $k \in \mathcal{K}$ $\&$ $l \in \mathcal{L}$. After these transformations, the original problem (P0) is approximated as a problem (P1) as follows:

    
\begin{eqnarray}        
   \text {\textbf{P1}}: & &  \max \limits _{\boldsymbol{{\Theta},\tilde{\mathbf{\Xi}}_k}}\quad \sum \limits ^{K}_{k=1} \mathbb {U}_{\alpha }(R_k), \label{eq:obj_fn} \\
     s.t.  & & (\ref{rate})~\&~(\ref{eq:constraint4}) \nonumber \\
    & &  \sum_{k=1}^{K} \sum_l^{{L}} \tilde{\xi}_{k,l} \leq {L}  \label{P1:constraint2},\\
    & &  \sum_{k=1}^{K}  \tilde{\xi}_{k,l} \leq 1, ~\forall l\in \mathcal{L} \label{P1:constraint3},\\
    & &  \tilde{\xi}_{k,l}  \in [0,1], ~\forall k \in \mathcal{K} ~\&~ l\in \mathcal{L}  \label{P1:constraint5}, 
\end{eqnarray}   
Constraints~(\ref{P1:constraint2}) and (\ref{P1:constraint3}) ensure that relaxed problem $\textbf{P1}$ still satisfies the original constraints (\ref{eq:constraint2}) and (\ref{eq:constraint3}).


Employing the BCD \cite{BCD} as in Algo. \ref{BCD} helps in finding a sub-optimal solution of \textbf{P1}. However, despite these improvements, the computational complexity of the BCD algorithm \cite{BCD} remains high, scaling as $O(K^2L^2+KL^2)$. This complexity poses challenges for practical implementation in real-time RIS-assisted networks. In the next section, we consider an unsupervised learning-based approach to address this limitation.
\section{Unsupervised learning-based approach}

We propose an unsupervised learning model to solve \textbf{P1}, where the objective is to jointly optimize the RIS phase shift ${\mathbf{\Theta}}$ and the allocation variables $  \tilde{\mathbf{\Xi}}_k$ in order to maximize the sum of the UEs’ utility rates.
\subsection{Input}
The input to the neural network is the channel state information (CSI), defined as: $ \mathbf{\mathcal{X}}=\{\mathbf{h}_k,{\mathbf{g}}_k,{\mathbf{H}}_{RB}\}_{k=1}^K$
, where $\mathbf{h}_k$ denotes the direct channel vector between the BS and the $k^{th}$ UE, $\mathbf{g}_k$ represents the channel between the RIS and the $k^{th}$ UE, and $\mathbf{H}_{RB}$ is the channel matrix between the BS and the RIS. Since these channel coefficients are complex-valued, the real and imaginary parts are separated before being used as input features. As a result, each data sample contains a total of $2 (LK + LN + KN)$ features.


\subsection{Fully connected Neural Network (FNN) Architecture} 
The proposed model is designed as a feedforward neural network comprising four fully connected layers, each followed by an activation function and a batch normalization layer. The fully connected layers perform linear transformations on the input features by applying a weight matrix and bias vector, thereby projecting the data into progressively higher-level representations. The activation functions are then applied element-wise, introducing non-linearity into the model and enabling it to capture complex, non-linear relationships in the data.

To further improve training stability and convergence, batch normalization is employed after each activation function. This operation normalizes the intermediate feature distributions to have zero mean and unit variance within each mini-batch, which mitigates internal covariate shift, accelerates training, and improves the overall generalization of the network.

The sequence of operations in each layer can thus be summarized as:
\begin{equation}
    \mathbf{c}^{(v)} = \text{BN} \left( \rho^{(v)}\left( \mathbf{\Omega}_w^{(v)} \mathbf{c}^{(v-1)} + \mathbf{\Omega}_b^{(v)} \right)\right) ,~\forall i \in \{1,\cdots,V\}
\end{equation}
where  $\mathbf{\Omega}_w^{(v)}$, $\mathbf{\Omega}_b^{(v)}$ and $\rho^{(v)}$ denote the weights, bias and activation function of the $v^{th}$ fully connected layer, respectively, and $\mathbf{c}^{(v)}$ represents the output of that layer. Also, $V$ denotes the number of hidden layers of the FNN. For each layer, the batch normalization operation is defined as:
\begin{equation}
    \text{BN}(c) =\frac{c - \mu_{BN}}{\sqrt{\sigma_{BN}^2 + \epsilon}}.
\end{equation}
Here, $\mu_{BN}$ and $\sigma_{BN}^2$ denote the mean and variance computed over the current batch input samples, $\epsilon$ is a small constant added for numerical stability.


\begin{algorithm}[t]
    \SetKwFunction{isOddNumber}{isOddNumber}
    \SetKwInOut{KwIn}{Input}
    \SetKwInOut{KwP}{Parameters}
    \SetKwInOut{KwOut}{Output}

    \KwIn{$\mathbf{\mathcal{X}}=\{\mathbf{h}_k,{\mathbf{g}}_k,{\mathbf{H}}_{RB}\}_{k=1}^K$}
    \KwP{$\mathbf{\Omega}$ - parameter vector of the NN, $\beta_1,\beta_2 \in [0,1)$ - decaying parameters, $\eta$-learning rate, $\epsilon$- a small constant for numerical stability}
    \KwOut{$\mathbf{\mathcal{Y}}=\{\mathbf{{\Theta}},\tilde{\mathbf{\Xi}}_k$\} }

    initialize parameter $\mathbf{\Omega}$
    
    $m_0,v_0\leftarrow 0$ 

    
    \For{$i \leftarrow 0$ \KwTo $n-1$}{
        $i \leftarrow i+1$
        
        Draw a random $Q$ samples from the $\mathbf{\mathcal{X}}$ : $\mathbf{\mathcal{X}}_Q = \{\mathbf{\mathcal{X}}_q\}_{q=1}^Q$

        Select only prominent features of input to reduce dimensionality, $f_d: \mathbf{\mathcal{X}}_Q \rightarrow \mathbf{\mathcal{Z}}_Q$ 
        
        
        
        
        $\mathbf{\mathcal{Y}}_Q = \text{NN}_{\mathbf{\Omega}}(\mathbf{\mathcal{Z}}_Q)$ 

        Compute the loss of the outputs, $L(\mathbf{\mathcal{Y}}_Q)$ as in (\ref{loss})

        
        $\mathbf{g}_i \leftarrow \nabla_\mathbf{\Omega}(L(\mathbf{\mathcal{Y}}_Q)$
        
        $\mathbf{m}_i \leftarrow \beta_1 . \mathbf{m}_{i-1} + (1- \beta_1) . \mathbf{g}_i$

        $\mathbf{v}_i \leftarrow \beta_2 .  \mathbf{v}_{i-1} +(1- \beta_2) . \mathbf{g}_i^2$

        ${\mathbf{\hat{m}}_i} \leftarrow \frac{\mathbf{m}_i}{1+\beta_1^i}$, $\mathbf{\hat{v}_i} \leftarrow \frac{{\mathbf{\hat{v}}_i}}{1+\beta_2^i}$
        
        $\mathbf{\Omega}_i \leftarrow \mathbf{\Omega}_{i-1} - \eta \frac{{\mathbf{\hat{m}}_i}}{\sqrt{{\mathbf{\hat{v}}_i}}+\epsilon}$
        

    }

    \KwRet{$\mathbf{\Omega}$}
    \caption{Unsupervised learning Algorithm}
    \label{UNN}
\end{algorithm}

\subsection{Output Layer} 
The output layers of the proposed neural network are designed to jointly estimate two outputs: $\mathbf{\mathcal{Y}} =\{\mathbf{{\Theta}},\tilde{\mathbf{\Xi}}_k$\} $=\text{NN}_{\mathbf{\Omega}}(\mathbf{\mathcal{X}})$, where $\mathbf{\Omega}=\{\mathbf{\Omega}_w,\mathbf{\Omega}_b\}$ denotes the trainable parameters of the neural network and $\mathbf{\mathcal{X}}$ represents the input feature vector. Specifically, the network is structured as follows:
\begin{itemize}
    \item The fifth fully connected layer is used to generate the estimated phase shifts of the RIS elements, denoted by $\mathbf{{\Theta}}$. This layer outputs continuous values that correspond to the phase configuration of the RIS.
    \item In parallel, another fully connected output layer followed by a sigmoid activation function produces the output variable $\tilde{\mathbf{\Xi}}_k$, which lies in the interval $[0,1]$. The sigmoid activation ensures that the estimated variable is bounded, which is often required when modeling probabilities, allocation coefficients, or other normalized quantities.
\end{itemize}
Once the outputs $\{\mathbf{{\Theta}},\tilde{\mathbf{\Xi}}_k\}$ are obtained, the training objective is defined through the following loss function:
\begin{equation}
L(\mathbf{\mathcal{Y}}_Q) = -\frac{1}{Q} \sum_{q=1}^{Q} \sum_{k=1}^{K} \mathbb{U}_{\alpha}(R_k),
\label{loss}
\end{equation}

where $Q$ denotes the number of training samples in a batch, and $K$ is the total number of UEs. The term $\mathbb{U}_{\alpha}(R_k)$ represents a utility function of the achievable rate $R_k$ for the $k^{th}$ UE. Thus, the network learns to output the optimal RIS phase shifts and resource allocation variables by minimizing the loss function in \eqref{loss} through backpropagation, whereby gradients of the loss are computed with respect to the network parameters and used to iteratively update them. This training process indirectly maximizes the desired system utility across all UEs in the batch. A schematic representation of the complete network architecture is provided in Fig.~\ref{CNN_archi}, illustrating the flow of data through the fully connected layers, activation functions, and batch normalization blocks.


\subsection{Data Preprossessing}
As the dimensionality of the input features increases, the number of parameters in a NN increases proportionally, particularly in the first layer where the parameter count scales linearly with the input size. This rapid growth not only escalates computational and memory demands, but also increases the risk of overfitting, since the model may memorize noise rather than learning meaningful patterns. To address this challenge, preprocessing techniques are employed. These techniques transform the original high-dimensional input into a more compact and informative representation, thereby reducing the effective input size while preserving the most discriminative information. Such preprocessing not only enhances computational efficiency but also strengthens the generalization capability of the model by minimizing redundancy, suppressing noise, and retaining only the most relevant features.

\begin{equation}
f_d : \mathbf{\mathcal{X}}  (\in \mathbb{R}^{D}) \rightarrow \mathbf{\mathcal{Z}} (\in \mathbb{R}^{D_{T}})
\end{equation}
where $D$ is the original feature size of the input and $D_{T}$ represents the reduced feature dimension after prepossessing. A brief description of the proposed neural network model that incorporates prepossessing step is illustrated in Fig.~\ref{CNN_archi}.




\section{ Numerical Results and Analysis}

\begin{table}[t]
\caption{Experimental Parameters description}
\begin{center}
\begin{tabular}{|c|c|}
\hline
\textbf{Simulation Parameters} & \textbf{Setup}\\
\hline
location of BS & (0m,0m,10m) \\
\hline
location of RIS & (25m,25m,10m) \\
\hline
number of antennas at BS $N$ & 4\\
\hline
density of UE $\lambda_k$ & 150$/km^2$\\
\hline
density of blockage $\lambda_b$&  $ 10/km^2$\\
\hline
number of RIS elements $L^2$ & $20 \times 20$ \\
\hline
center frequency $f_c$ & 28GHz \\
\hline 
Small scale shadow fading $\sigma_\text{SF}$ & 4\\
\hline
noise level $\sigma^2$ &\ -84dBm\\
\hline

transmitter power $P_t$ & 35dBm \\
\hline

decaying parameters $\beta_1, \beta_2$ & 0.9, 0.999\\
\hline

learning rate $\eta$ & 0.01 \\
\hline

Batch size & 20\\

\hline

numerical stability constant $\epsilon$ & $10^{-8}$\\
\hline

Bandwidth $B$ & 50MHz\\

\hline

\end{tabular}
\label{tab1}
\end{center}
\end{table}
We consider a single RIS-aided MIMO wireless network, illustrated in Fig \ref{System Model}, to establish the simulation environment in Python. The experiments are carried out on a workstation with an Intel i9-13900K processor and NVIDIA GeForce RTX 2080 Ti graphics processor of 12GB. For the simulations, we consider an area of $100 \times 100~\text{m}^2$. A single BS equipped with four antennas and a RIS comprising $20 \times 20$ elements are deployed at fixed locations $(0\text{m}, 0\text{m}, 10\text{m})$ and $(25\text{m}, 25\text{m}, 10\text{m})$, respectively. The BS and RIS are assumed to be arranged in a Uniform Planar Array (UPA) configuration, lying on the xz-plane and xy-plane, respectively. The UEs are randomly distributed on the same side of both the BS and RIS, with a density of $150/\text{km}^2$. Blockages are modeled with an intensity of $5/\text{km}^2$, where each blockage has an average length and width of $15\text{m}$. The carrier frequency and bandwidth are set to $28 \text{GHz}$ and $50 \text{MHz}$, respectively. We also account for small-scale fading with a standard deviation of $\sigma_\text{SF} = 4$. Furthermore, we assume that the beamforming vectors satisfy $\lVert \mathbf{w}_k \rVert^2 = P_t/K, ; \forall k \in \mathcal{K}$, which does not affect the optimization problem \textbf{P1} or \textbf{P0}. This assumption ensures equal power allocation among UEs while simplifying the system design. Since the objective of this work is to investigate RIS phase and element allocation, transmit beamforming is kept fixed to avoid introducing additional optimization variables and complexity. Such a setting is consistent with prior studies in RIS-aided systems \cite{fixed_beamforming}, where fixing $\mathbf{w}_k$ allows isolating and clearly evaluating the performance gains attributable to the RIS.

The described simulation environment is implemented in accordance with the 3GPP specifications \cite{3GPP}. Using this setup, we generate $8,000$ data samples for training and $2,000$ data samples for validation. The training dataset is used to optimize the parameters of the neural network, while the validation dataset is employed for tuning the hyperparameters, such as the depth and width of the FNN.

\begin{table}[t]
\caption{Architecture of the proposed FNN with prepossessing as PCA}
\begin{center}
\begin{tabular}{|c|c|}
\hline
\textbf{Layer} & \textbf{Output Shape}\\
\hline
Input & (10000, 15335) \\
\hline
PCA & (15335,6)\\
\hline
fully connected  layer1 & (6,500) \\
\hline
ReLU 1 & (1655,500) \\
\hline
Batch Normalization 1 & (1655,500) \\
\hline
fully connected layer 2 & (500,450) \\
\hline
ReLU 2 & (500,450) \\
\hline
Batch Normalization 2 & (500,450) \\
\hline
fully connected layer 3 & (450,400) \\
\hline
ReLU 3 & (450,400) \\
\hline
Batch Normalization 3 & (450,400) \\
\hline
fully connected layer 4 & (400,300) \\
\hline
ReLU 4 & (400,300) \\
\hline
Batch Normalization 4 & (400,300) \\

\hline
fully connected layer \Romannum{1} & (300, 800) \\
\hline
fully connected layer \Romannum{2} & (300, 12000) \\
\hline
Sigmoid layer \Romannum{1} & (300, 12000) \\
\hline

\end{tabular}
\label{FNN_archi}
\end{center}
\end{table}


\begin{figure}[t]
\vspace{-0.275in}
\centerline{\includegraphics[scale=0.5]{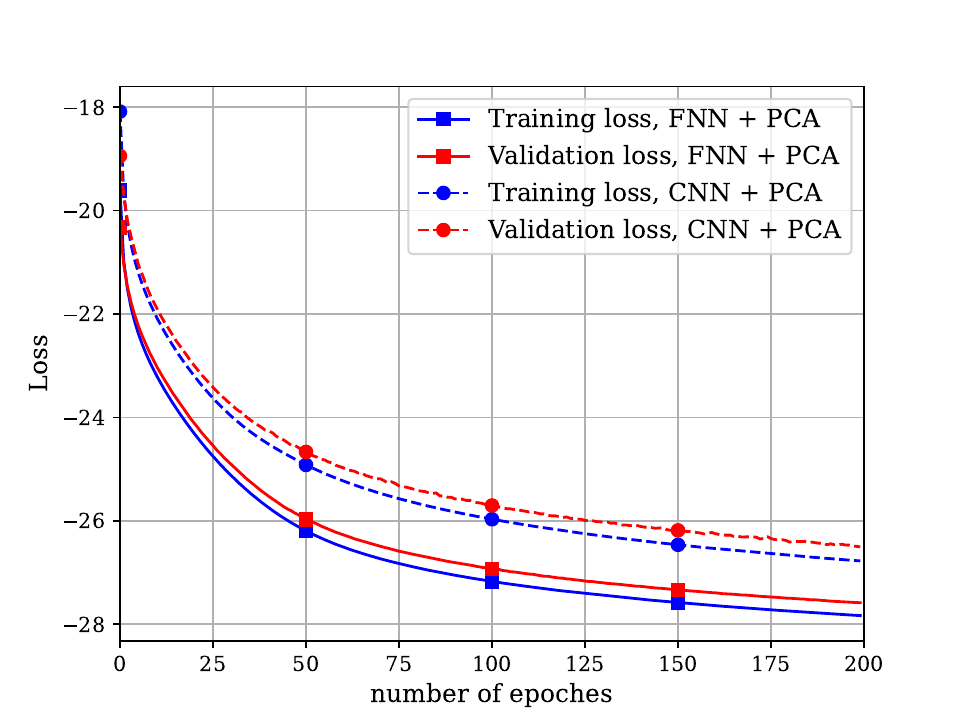}}
\caption{Training and validation losses of the CNN and FNN for the proposed RIS element allocation framework with fairness parameter $\alpha = 1$, RIS size $L^2 = 20 \times 20$, and six PCA components.}
\label{Proposed RIS RA}
\end{figure}

\begin{table*}
\centering
\caption{Performance comparison of existing and proposed RIS element allocation schemes employing FNN and PCA for RIS configurations of $L^2 = 40 \times 40$ and $20 \times 20$}
\label{PerfComp}
\begin{tabularx}{\textwidth}{@{} l *{6}{C} c @{}}
\hline
\multicolumn{1}{c}{} & \multicolumn{3}{c}{$L^2=40 \times 40$} & \multicolumn{3}{c}{$L^2=20 \times 20$ } \\ \cline{2-4} \cline {5-7}
\multirow{3}*{} &\multicolumn{2}{c}{Avg. } & Network    &\multicolumn{2}{c}{Avg. } & Network \\
&\multicolumn{2}{c}{$\alpha$-mean Throughput } & Parameters  &\multicolumn{2}{c}{$\alpha$-mean Throughput } & Parameters   \\
&\multicolumn{2}{c}{( $\times10^8$ bps)}  &   &\multicolumn{2}{c}{( $\times10^8$ bps)} &    \\

\cline{2-3} \cline{5-6}
\multicolumn{1}{c}{} & \multicolumn{1}{c}{$\alpha=1$} & \multicolumn{1}{c}{$\alpha=2$} & \multicolumn{1}{c}{} & \multicolumn{1}{c}{$\alpha=1$} & \multicolumn{1}{c}{$\alpha=2$} & \multicolumn{1}{c}{}&\\
\hline
Existing RIS element allocation \cite{5b} &  3.448 & 3.447 & 10.564M  & 3.441 & 3.441 & 2.960M \\ \hline
Proposed RIS element allocation & 3.274 & 3.273 & 39.184M  & 3.176 & 3.176 & 10.264M \\ \hline

Existing RIS element allocation \cite{5b} + PCA & 3.153 & 3.152 & 0.742M  & 2.823 & 2.824 & 0.546M \\ \hline

\textbf{Proposed RIS element allocation + PCA} &  \textbf{3.752} & \textbf{3.745} & \textbf{0.962M}  & \textbf{3.695} & \textbf{3.694} & \textbf{0.148M}  \\ \hline
\end{tabularx}
\end{table*}

The unsupervised learning models are trained using the Adam optimizer \cite{adam} with an initial learning rate of $0.01$ and a batch size of $20$. To accelerate convergence in the early training stages and reduce oscillations in later stages, a learning rate decay strategy is adopted, where the learning rate is reduced by a factor of $0.33$ if no improvement is observed over ten consecutive epochs. To mitigate overfitting, early stopping with a patience of $40$ epochs is employed, along with dropout regularization at a rate of $0.5$.

The complete neural network architecture, summarized in Table~\ref{FNN_archi}, is carefully designed to provide a good fit for the optimization task. In the hidden layers, the Rectified Linear Unit (ReLU) activation function, $\text{ReLU}(x)=\max(0,x)$, introduces non-linearity, enabling the model to learn complex feature representations. Batch normalization (BN) is applied after each hidden layer to stabilize training and improve generalization. The output layer consists of two parallel fully connected layers (layers \Romannum{1} and \Romannum{2}), with one layer followed by a sigmoid activation function for probabilistic resource allocation output.
Furthermore, principal component analysis (PCA)~\cite{PCA} is employed as a preprocessing step to project the original high-dimensional feature space into a lower-dimensional subspace while retaining the directions of maximum variance. The number of retained components is determined using the Kaiser rule~\cite{Kaiser}, which selects only those with eigenvalues greater than unity. This ensures that each retained principal component explains more variance than any single original feature, thereby improving computational efficiency and model generalization while reducing redundancy and noise in the input.

The training and validation datasets are used to train the neural network, as summarized in Table~\ref{FNN_archi}. Training and validation losses of the CNN and FNN for fairness parameter $\alpha=1$, under the proposed RIS element allocation framework with $20 \times 20$ RIS elements and six PCA components (see Fig.~\ref{Proposed RIS RA}).
 The loss curves of the FNN, shown in Fig.~\ref{Proposed RIS RA}, demonstrate a good model fit, as the validation loss remains only slightly higher than the training loss. A good fit ensures that the neural network learns significant patterns from the training data while maintaining strong generalization to unseen data. Furthermore, for benchmarking, a convolutional neural network (CNN)\cite{CNN} is also trained under the same setup. As illustrated in Fig.\ref{Proposed RIS RA}, the FNN consistently outperforms the CNN in addressing the transformed problem \textbf{P1}. In contrast to CNNs, which rely on local receptive fields that are less appropriate for the structured, high-dimensional data in this task, the FNN's fully connected layers efficiently capture global feature dependencies, which is why it performs better. This outcome highlights the effectiveness of the FNN architecture for the proposed RIS-based optimization problem.

  \begin{table}[t]
\centering
\caption{Average Sum rate comparison of existing and proposed RIS element allocation schemes for $L^2 = 40 \times 40$ and $20 \times 20$ RIS configurations}
\label{SumRateComp}
\begin{tabular}{lcc}
\hline
\textbf{} & \textbf{$40 \times 40$} & \textbf{ $20 \times 20$} \\
\hline
Existing RIS element allocation \cite{5b} & 103.421 & 103.249 \\
\hline
Proposed RIS element allocation & 98.213 & 95.293 \\
\hline
Existing RIS element allocation \cite{5b} + PCA & 94.589 & 84.718 \\
\hline
\textbf{Proposed RIS element allocation + PCA} & \textbf{112.335} & \textbf{95.348} \\
\hline
\end{tabular}
\end{table}

\subsection{Benchmarking with Baseline Models}

The results presented in Table \ref{PerfComp} provide a comparative analysis of the proposed RIS element allocation framework against existing baselines for two RIS sizes, $L^2= 40\times40$ and $L^2= 20\times20$. The evaluation is conducted using the average $\alpha$-mean throughput ($\alpha= 1, 2$) as a measure of system performance, alongside the number of trainable network parameters to capture computational complexity.

\begin{table}[t]
\centering
\caption{$\%$ Bandwidth saving  comparison of existing and proposed RIS element allocation schemes for $L^2 = 40 \times 40$ and $20 \times 20$ RIS configurations}
\label{BndwitdthSavingComp}
\begin{tabular}{lcc}
\hline
\textbf{} & \textbf{$40 \times 40$} & \textbf{ $20 \times 20$} \\
\hline
Existing RIS element allocation \cite{5b} & 97.099 & 97.094\\
\hline
Proposed RIS element allocation & 96.941 & 96.852 \\
\hline
Existing RIS element allocation \cite{5b} + PCA & 96.828 & 96.458 \\
\hline
\textbf{Proposed RIS element allocation + PCA} & \textbf{97.329} & \textbf{96.877} \\
\hline
\end{tabular}
\end{table}

For the larger RIS configuration ($40 \times 40$), the existing RIS element allocation achieves a stable throughput of around $3.44 \times 10^8$bps with $10.564\text{M}$ parameters. The proposed allocation without PCA, although functional, significantly inflates the complexity of the model to the $39.184\text{M}$ parameters, while it does not offer substantial gain in throughput, revealing limitations in scalability. When PCA is applied, the existing method with PCA reduces complexity to below $1\text{M}$ parameters but suffers a noticeable drop in average throughput to $3.15 \times 10^8$bps. In contrast, the proposed method with PCA strikes an optimal balance by achieving the highest average throughput of $3.752 \times 10^8$bps while keeping the complexity low at just $0.962\text{M}$ parameters. A similar trend is observed in the smaller RIS case ($20 \times 20$). The baseline method requires $2.960\text{M}$ parameters to deliver $3.441 \times 10^8$bps, while the proposed non-PCA model consumes $10.264\text{M}$ parameters for similar performance. With PCA, the existing method compresses the parameters to $0.546\text{M}$ but sacrifices throughput, which falls to $2.824 \times 10^8$bps. However, the proposed PCA-based model achieves both objectives: it yields the best performance throughput with minimal complexity, representing a substantial efficiency gain, as highlighted in Table~\ref{PerfComp}.

In summary, the analysis of Table \ref{PerfComp} shows that the proposed RIS element allocation combined with PCA delivers the best performance–efficiency trade-off. It achieves the highest throughput across both RIS configurations, while using orders of magnitude fewer parameters compared to conventional approaches. This clearly validates the importance of incorporating feature reduction (via PCA) into RIS allocation, making the framework both computationally scalable and throughput-efficient for practical large-scale deployments.

 To support our claim, we also compare the average sum rate and $\%$ bandwidth saving, as presented in Tables~\ref{SumRateComp} and ~\ref{BndwitdthSavingComp}. The sum rate analysis confirms that the proposed RIS element allocation with PCA consistently outperforms conventional schemes for the $L^2 = 40 \times 40$ and $20 \times 20$ RIS configurations, achieving higher spectral efficiency while requiring significantly fewer network parameters. Similarly, the bandwidth saving analysis highlights that the proposed framework achieves substantial reductions in spectrum usage, indicating more efficient utilization of wireless resources. Together, these results reinforce the effectiveness of integrating FNN with PCA, demonstrating that the framework not only reduces computational burden but also provides superior system performance and spectrum efficiency.

\begin{table}[t]
    \centering
    \caption{ Performance comparison of Conventional iterative optimization and Unsupervised FNN for $40 \times 40$ RIS elements}
        \begin{tabular}{cccc}
        \hline
        \multirow{3}*{} &\multicolumn{2}{c}{Avg. } & Inference  \\
        &\multicolumn{2}{c}{$\alpha$-mean Throughput }  & time  \\
        &\multicolumn{2}{c}{( $\times10^8$ bps)}  & (in minutes)  \\
        \cline{2-3} 
        \multicolumn{1}{c}{} & \multicolumn{1}{c}{$\alpha=1$} & \multicolumn{1}{c}{$\alpha=2$} &  \\
         \hline
         BCD & 74.323 & 72.365 & 200 \\ 
         \hline
         \textbf{FNN + PCA} & \textbf{112.231}& \textbf{112.230}  & \textbf{$<$1} \\ 
         \hline
    \end{tabular}
\label{BCDvsFNN}
\end{table}

Furthemore, we compare the average $\alpha$-mean throughput (with $\alpha = 1, 2$) and the inference time, defined as the time required to generate outputs in real-time applications, of the proposed RIS element allocation using the conventional BCD algorithm and the unsupervised learning-based FNN integrated with PCA. The performance comparison for $40 \times 40$ RIS elements is summarized in Table~\ref{BCDvsFNN}.

The results clearly demonstrate the superiority of the FNN with PCA over the BCD algorithm. In terms of throughput, the FNN consistently achieves higher $\alpha$-mean values, indicating more efficient utilization of RIS elements and better adaptation to user requirements. This improvement stems from the FNN’s ability to capture complex, non-linear mappings between input features and optimal RIS configurations, a capability that iterative optimization in BCD fails to exploit fully.

With respect to inference efficiency, the BCD algorithm, owing to its iterative nature, becomes computationally intensive and time-consuming as the system size increases. This makes it impractical for real-time operation, especially in large-scale RIS deployments where $K$, $L$, and $N$ are large. In contrast, once trained, the FNN is able to generate inference results in under a minute for the considered RIS configuration, ensuring responsiveness suitable for real-time wireless systems.

\subsection{Computational Complexity Analysis}
The computational complexity of the algorithms considered is summarized as follows. For the BCD benchmark algorithm, the complexity is $O\left(K^{2}L^{2} + KL^{2}\right),$ which grows quadratically with the problem dimensions $K$ and $L$. In contrast, FNN based on unsupervised learning has a complexity of  
\begin{align}
O\left(\sum_{v=1}^{V} c_{v-1} c_{v}\right) \approx O(KL + KN + LN),
\label{FNN_CC}
\end{align}
where $V$ is the depth of the network, $c_0$ denotes the input dimension, and $c_v$ represents the number of neurons in the $v^{th}$ hidden layer. This shows that the FNN scales linearly with the system parameters, resulting in a significant reduction in computational burden compared to BCD. Furthermore, when dimensionality reduction techniques such as PCA are applied, the effective input dimension is reduced to $D_T$, where $D_T \ll (KL + KN + LN)$. Consequently, the overall complexity becomes $O(D_T)$, further reducing the training cost. Beyond computational efficiency, PCA also enhances generalization by mitigating overfitting and removing redundancy, making the learning process both faster and more robust.

As evidenced in Table~\ref{BCDvsFNN}, the FNN with PCA not only provides dramatic gains in computational efficiency and inference speed, but also surpasses BCD in terms of throughput performance. This dual advantage establishes the FNN + PCA framework as a highly effective solution for large-scale RIS optimization tasks. It strikes an excellent balance between efficiency, scalability, and accuracy, making it especially well-suited for real-time deployment in next-generation RIS-aided wireless systems, where both performance and responsiveness are critical.

\section{Conclusion}

In this work, we proposed a discrete RIS element allocation scheme using an unsupervised learning-based approach for multiple UEs in mmWave networks. Unlike conventional iterative optimization methods, which suffer from exponentially increasing complexity and the difficulty of generating training labels, the proposed framework employs a five-layer fully connected neural network to achieve scalable, efficient, and fair RIS allocation without relying on supervised data. The simulation results show that the proposed method reduces computational overhead while significantly improving system throughput and fairness. Specifically, it achieves a throughput gain of 6.8$\%$ compared to existing RIS allocation methods~\cite{5b} and 22$\%$ compared to conventional iterative optimization techniques~\cite{BCD}. Furthermore, with PCA-based preprocessing, the framework achieves higher spectral efficiency with substantially fewer parameters, ensuring an excellent performance–efficiency trade-off.

The results also reveal that uniform and contiguous RIS allocation may restrict system performance, whereas the proposed discrete allocation strategy provides greater flexibility and fairness. Overall, the results establish unsupervised neural network-based RIS allocation as a promising solution for scalable and practical real-time deployment in RIS-assisted mmWave systems, while future work will address imperfect CSI, dynamic user mobility, and hardware constraints to further enhance its applicability in next-generation wireless networks.

\bibliographystyle{IEEEtran}
\bibliography{IEEEabrv,citations}

\end{document}